\def\year{2020}\relax
\documentclass[letterpaper]{article} % DO NOT CHANGE THIS
\usepackage{aaai20}  % DO NOT CHANGE THIS
\usepackage{times}  % DO NOT CHANGE THIS
\usepackage{helvet} % DO NOT CHANGE THIS
\usepackage{courier}  % DO NOT CHANGE THIS
\usepackage[hyphens]{url}  % DO NOT CHANGE THIS
\usepackage{graphicx} % DO NOT CHANGE THIS
\usepackage{graphicx}  % DO NOT CHANGE THIS
\usepackage{booktabs}
\usepackage{makecell}
\usepackage{subcaption}
\usepackage{xcolor}
\usepackage{pgfplotstable, tabularx}
\usepackage{threeparttable}
\usepackage{amsfonts}
\usepackage{amsmath}
\usepackage{numprint}
\newcommand*{\fnref}[1]{\textsuperscript{\ref{#1}}}
\title{
Empirical Analysis of Multi-Task Learning for Reducing Identity Bias \\in Toxic Comment Detection
}

   \author{Ameya Vaidya,\textsuperscript{1} Feng Mai,\textsuperscript{2} Yue Ning\textsuperscript{3} \\ 
  \textsuperscript{1}{Bridgewater-Raritan Regional High School}\\
  \textsuperscript{2}{School of Business, Stevens Institute of Technology}\\
  \textsuperscript{3}{Department of Computer Science, Stevens Institute of Technology}\\
  ameyav993@gmail.com, fmai@stevens.edu, yue.ning@stevens.edu % email address must be in %roman text type, not monospace or %sans serif
  }
 
\begin{document}

\maketitle

\begin{abstract}

With the recent rise of toxicity in online conversations on social media platforms, using modern machine learning algorithms for toxic comment detection has become a central focus of many online applications. Researchers and companies have developed a variety of models to identify toxicity in online conversations, reviews, or comments with mixed successes. However, many existing approaches have learned to incorrectly associate non-toxic comments that have certain trigger-words (e.g. gay, lesbian, black, muslim) as a potential source of toxicity. In this paper, we evaluate several state-of-the-art models with the specific focus of reducing model bias towards these commonly-attacked identity groups. We propose a multi-task learning model with an attention layer that jointly learns to predict the toxicity of a comment as well as the identities present in the comments in order to reduce this bias. We then compare our model to an array of shallow and deep-learning models using metrics designed especially to test for unintended model bias within these identity groups. 

\end{abstract}

\section{Introduction}

The identification of potential toxicity within online conversations has always been a significant task for current platform providers. Toxic comments have the unfortunate effect of causing users to leave a discussion or give up sharing their perspective and can give a bad reputation to platforms where these discussions take place. Twitter's CEO reaffirmed that Twitter is still being overrun by spam, abuse, and misinformation~\cite{twitterceo}.
% \footnote{\url{https://tinyurl.com/y8x3xvv2}} 
To deal with this problem, researchers and companies have done extensive investigations into the field of toxic comment detection. Current research involves tackling common challenges in toxic comment classification~\cite{van-aken-etal-2018-challenges}, identifying subtle forms of toxicity~\cite{Noever2018MachineLS}, detecting early signs of toxicity~\cite{zhang:2018}, and analysing sarcasm within conversations~\cite{ghosh-etal-2018-sarcasm}. 

Over the past few years, a variety of models and methods have been proposed to detect online toxic comments. Current baseline methods exploit the representation of documents as character n-grams or TF-IDF~\cite{Badjatiya:2017:DLH} which are then learned by Logistic Regression or Support Vector Machines~\cite{Noever2018MachineLS}. Recently, deep learning methods such as convolutional neural networks~\cite{Georgakopoulos:2018:CNN} and recurrent neural networks~\cite{zhou-etal-2016-text} have been popularized in natural language processing to analyze online content. Furthermore, bidirectionality~\cite{zhou-etal-2016-text}, attention mechanisms~\cite{attention:2015}, and ensemble learning~\cite{Dietterich:2000:EMM} have also shown improved performance in text sentiment analysis. 
\begin{table}[t]
    \centering
    \small
      \caption{Example of toxic comments with identity attack where Identity can be replaced by ``gay'', ``black'' etc.}
    \label{tab:example}
    \begin{tabular}{cp{3cm}p{3cm}}
    \toprule
        \textbf{Identity} &  \textbf{Toxic(\textcolor{red}{\textbf{+}})} & \textbf{Non-toxic(\textcolor{green}{\textbf{-}})}\\
        \hline
         Positive  & \textit{$\langle$\colorbox{gray!10}{Identity}$\rangle$ people are gross and universally hated!} & \textit{I am a $\langle$\colorbox{gray!10}{Identity}$\rangle$ person, ask me anything.}\\
         \hline
        Negative & \textit{What the heck is wrong with you?} & \textit{Thanks for the help. I really appreciate it!}\\
        \bottomrule
    \end{tabular}
\end{table}

However, many existing works have documented that current toxic comment classification models introduce bias into their predictions. They tend to classify comments that reference certain commonly-attacked identities (e.g., gay, black, muslim) as toxic without the comment having any intention of being toxic~\cite{Dixon:2018:MMU,Borkan:2019:NMM} as shown in Table~\ref{tab:example}. For example, the comment ``I am a black woman, how can I help?'' might be classified by a model as toxic because it references the `black' identity. Furthermore, the Conversation AI team at Google Jigsaw has acknowledged that their Perspective API framework, which attempts to detect toxicity in online conversations, seems to generate higher toxicity scores for sentences containing commonly targeted identity groups.\footnote{\url{https://medium.com/the-false-positive/unintended-bias-and-names-of-frequently-targeted-groups-8e0b81f80a23}} 
Current research efforts to investigate model bias~\cite{sap-etal-2019-risk,davidson-etal-2019-racial} have detected a correlation between race, identity, and model predictions in the context of hate and abusive speech. Evaluation metrics have also been developed to test specifically for implicit model bias~\cite{Dixon:2018:MMU}. However, not many novel mitigation solutions have been proposed within current research efforts, which is something that we hope to contribute with the proposal of our model.

In this paper, our main focus is to reduce the false positive rate on non-toxic comments that make reference to identities known historically and empirically to introduce model bias. Our empirical analysis focuses specifically on the improvement of the following identities because of their tendency to be associated with a high false positive rate: gay, lesbian, bisexual, transgender, black, muslim, and jewish. To deal with this challenge, we propose a multi-task learning framework that simultaneously identifies toxicity and identity information within a comment. 
Learning these tasks jointly will allow the model to share common patterns and better distinguish between toxic and non-toxic comments that reference these identities. 
This paper also aims to evaluate various shallow and deep learning models adapted from existing literature, including logistic regression and recurrent neural networks, on the task of mitigating bias. We evaluate the proposed multi-task learning model and other deep learning methods on a dataset of 1,804,874 unique comments published by Google Jigsaw during Kaggle's Unintended Bias in Toxicity Classification Challenge.\footnote{\label{jigsaw}\url{https://www.kaggle.com/c/jigsaw-unintended-bias-in-toxicity-classification/overview}} To keep our focus on mitigating unintended bias, we utilize a set of evaluation metrics that are specifically designed for measuring bias in the model outputs.

The ultimate goal of our research is to help maintain the civility of conversations on common social media platforms while minimizing the amount of non-toxic comments that are classified as toxic. 
Our main contributions are summarized as below:

\begin{enumerate}
\item We perform an empirical study for a multitude of classifiers on a new public dataset containing over 1.8 million comments. We also compare classifiers with the specific focus of reducing unintended model bias within online conversations.
\item We propose a multi-task learning model that outperforms other models at mitigating unintended bias, especially for certain identities that historically bring a high rate of false positives. The attention layer included in the multi-task learning model allows us to capture hidden state dependencies and distinguish between toxic and non-toxic comments. We also employ a custom-weighted loss function that allows our model to increase penalization on false positive mistakes.
\item We analyse the classifiers' predictions on a variety of evaluation metrics. These measures include F1-measures and the AUC-ROC score. In addition, we evaluate our models on metrics designed specifically for unintended model bias: Generalized Mean Bias AUC, Subgroup AUC, and BPSN AUC. We also compare non-toxic and toxic comments across models and with Google's Perspective API.
\end{enumerate}

In the following sections, we will introduce the related work and the investigated dataset followed by the proposed multi-task learning framework. Then we discuss the experimental evaluation and results. Finally we conclude our work with future directions.

\section{Related Work}
In this section, we will briefly review recent developments in multi-task learning. We will then focus on new attempts on toxic comment classification and identity bias in natural language processing models.

\subsection{Multi-Task Learning}
Multi-task learning~\cite{Caruana1998,mtl:2007} has been widely studied and applied in natural language processing (NLP)~\cite{Collobert:2008:UAN,deng:2013}, computer vision, and other machine learning applications~\cite{Ramsundar2015MassivelyMN}. In deep learning models, multi-task learning is usually implemented by either sharing hidden layer model parameters~\cite{Long:2017:LMT} or regularizing parameters among related tasks to be similar~\cite{duong-etal-2015-low}. Recent works show that multi-task learning can improve performance on various NLP tasks while revealing novel insights about language modeling~\cite{sogaard2016deep}. In terms of network architecture, our work is closest to the LSTM-based multi-task learning frameworks \cite{liang-shu-2017-deep,Suresh:2018:LTM}. However it is known that the performance of multi-task learning is task specific~\cite{misra2016cross}. Which framework is more effective at teasing out identity information while detecting toxic comments is an open empirical question. 

\subsection{Toxic Comment Detection}

Machine learning for detecting toxic comments has been a significant focus in Natural Language Processing research over the past few years. This is in part due to the availability of large corpora of online social interactions. Wikimedia Foundation \cite{wulczyn2017ex} released an annotated dataset of personal attacks, toxic messages, and aggression from the English Wikipedia Talk pages. Google Jigsaw also published two Kaggle competitions which have allowed researchers to gain access to datasets with 2.5 million training examples of toxic comments.
In terms of methods, most research takes a text classification approach similar to sentiment analysis and spam detection \cite{mishra2018author,davidson2017automated,wulczyn2017ex}. These methods rely on document features (readability, emotion, sentiment, n-grams), author features (demographics, social network positions), or contextual features (the relationship of a document to others) to train classifiers. 

More recent research advances toxic comment detection models on two fronts. Some studies move beyond using documents as the units of analyses and model the behavior of the users~\cite{cheng2015antisocial}, or take a more proactive approach to detect online conversations that are susceptible to escalation~\cite{zhang:2018}. Another stream of work uses neural network models to classify toxic comments and has shown impressive results~\cite{Georgakopoulos:2018:CNN,chen2019use,elnaggar:18,srivastava-etal-2018-identifying}. Although these new models can achieve good performance without hand-crafted features, a potential downside is that the decisions made by the classifiers are more opaque. When the model is deployed, it may conflate identity attacks with identity disclosures, and make a biased decisions against the latter. Our work extends this stream of research and uses multi-task learning to explicitly account for the identity bias. 

\subsection{Unintended Identity Bias in NLP Models}
A growing number of studies have called attention to the identity related biases in natural language models. Several studies have highlighted how word embeddings exhibit human stereotypes towards genders and ethnic groups ~\cite{bolukbasi2016man,caliskan2017semantics,garg2018word}. One way to counter such biases is to enhance the interpretability of black box models~\cite{guidotti2019survey} so that humans can intervene when a model makes an unfair decision. Another way to address the issue, which is the focus of this study, is to design models to circumvent protected identity attributes. Several methods have been proposed in the context of structured or numerical data ~\cite{corbett2018measure}, but methods applicable to text data generated by online users are rare. During our research, we discovered the Pinned AUC metric which is popularly used to mitigate unintended bias~\cite{Dixon:2018:MMU}. However, we have decided its use is unwarranted in this paper due to recent discoveries which suggest that Pinned AUC can be distorted by uneven distributions, which is prevalent within the dataset we analyze~\cite{Borkan2019LimitationsOP}.

\section{Dataset}
\begin{figure}[t]
\centering
  \includegraphics[width=0.8\linewidth]{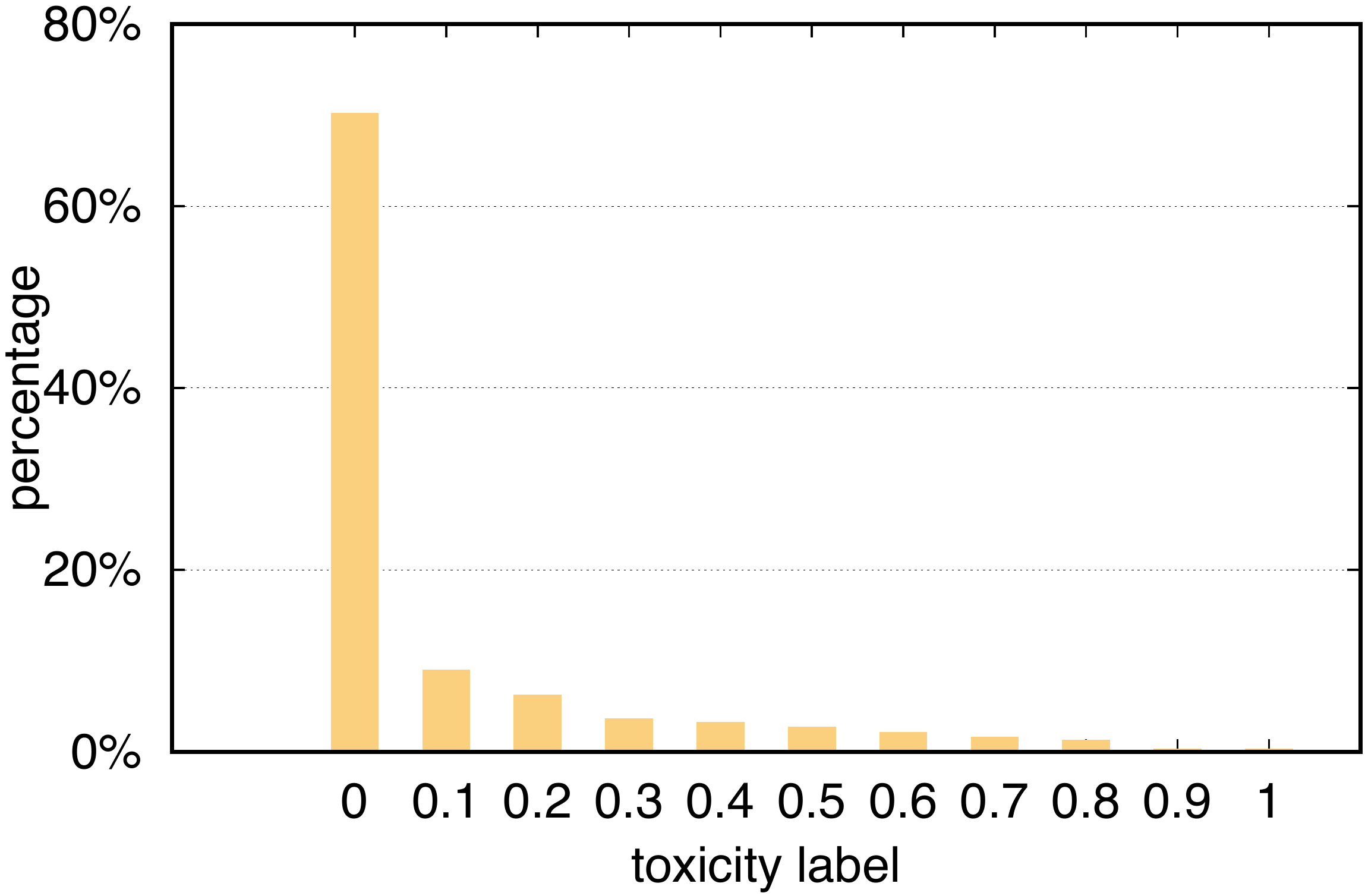}
  \caption{Percentage distribution of toxicity labels in the dataset. X-axis is the interval of toxicity scores (e.g. `0'=[0,0.1)) and Y-axis is the combined percentages of comments in each interval. It shows a clear imbalanced distribution of toxic and non-toxic comments in the dataset. }
  \label{fig:target_dist}
\end{figure}

\noindent We analyse a dataset published by the Jigsaw Unintended Bias in Toxicity Classification Challenge on Kaggle. It contains 1,804,874 comments annotated by the Civil Comments platform. Each comment is shown up to 10 annotators who classify each comment as either \textit{Very Toxic}, \textit{Toxic}, \textit{Hard To Say}, or \textit{Not Toxic}. Each comment is then given a toxicity label based on the fraction of annotators that classified it as either \textit{Toxic} or \textit{Very Toxic}. For evaluation, every comment with a toxicity label greater than or equal to 0.5 was considered to be toxic (the positive class). As discussed by Jigsaw, a toxic comment usually contains rude, disrespectful, or unreasonable content that is somewhat likely to make you leave a discussion or give up on sharing your perspective.
\pgfplotstableset{
    every head row/.style={
    before row=\toprule,after row=\midrule},
    every last row/.style={
    after row=\bottomrule},
    col sep = &,
    row sep=\\,
    string type,
}
\begin{table}
\caption{Number of comments labeled with each identity and percent of comments in each identity that are non-toxic.}\label{tab:data}
\begin{tabular}{lcc}
\toprule
\textbf{Identities}                          & \textbf{Count}      &\textbf{Non-Toxic}      \\ 
\hline
male                                & 64,544     & 90.40\%        \\
female                              & 55,048     & 90.86\%        \\
homosexual (gay or lesbian)         & 11,060     & 80.99\%        \\
christian                           & 40,697     & 94.41\%        \\ 
muslim                              & 21,323     & 85.06\%        \\
jewish                              & 7,669      & 89.75\%        \\
black                               & 17,161     & 80.31\%        \\
white                               & 28,831     & 82.24\%        \\
psychiatric or mental illness       & 6,218      & 85.91\%        \\
All Identities                      & 191,671    & 85.56\%        \\
\bottomrule
\end{tabular}
% }
\end{table}

In addition, each comment was also labeled with a multitude of identity attributes (non-exclusive), which demonstrates the presence of a specific identity in a comment. These identities include \textit{male}, \textit{female}, \textit{homosexual (gay or lesbian)}, \textit{christian}, \textit{jewish}, \textit{muslim}, \textit{black}, \textit{white}, and \textit{psychiatric or mental illness}. Label values were given based on the fraction of annotators who believed a comment fit the identity mentioned. Each comment was also labeled with five subtype attributes: \textit{severe\_toxicity}, \textit{obscene}, \textit{threat}, \textit{identity\_attack}, and \textit{insult} based on the percent of annotators who identified a comment with the aformentioned subtype. Out of these nine identities, we found that the following identities tended to have the highest false positive rates: \textit{homosexual}, \textit{muslim}, \textit{jewish}, \textit{black}.

Figure~\ref{fig:target_dist} suggests that the distribution of the toxicity label within the dataset follows a long tail distribution. Approximately 92\% of the comments are classified as non-toxic (negative class). Table~\ref{tab:data} shows the distribution of identity labels in the dataset. In addition, within this dataset, the average document length is approximately 51.28 words long, meaning that identifying long-range dependencies is an important consideration in this paper. We discuss the use of Long Short-Term Memory Networks (LSTM) later in this paper to deal with this challenge.

\section{Models and Tasks}

In this section, we explore and propose a multi-task learning framework whose focus is to improve the accuracy of correctly detecting toxic comments by jointly learning toxicity and identity information. The toxicity task aims to correctly predict the toxicity score for a comment. The identity task is designed to predict the presence of an identity in a comment. These tasks work jointly to reduce the model bias towards commonly attacked identities in Table~\ref{tab:data}. 
% \textcolor{red}{The code is open sourced on Github: \url{https://github.com/}}.
% This model was implemented using the Keras python framework.

\begin{figure}[t]
\centering
\includegraphics[width=0.35\textwidth]{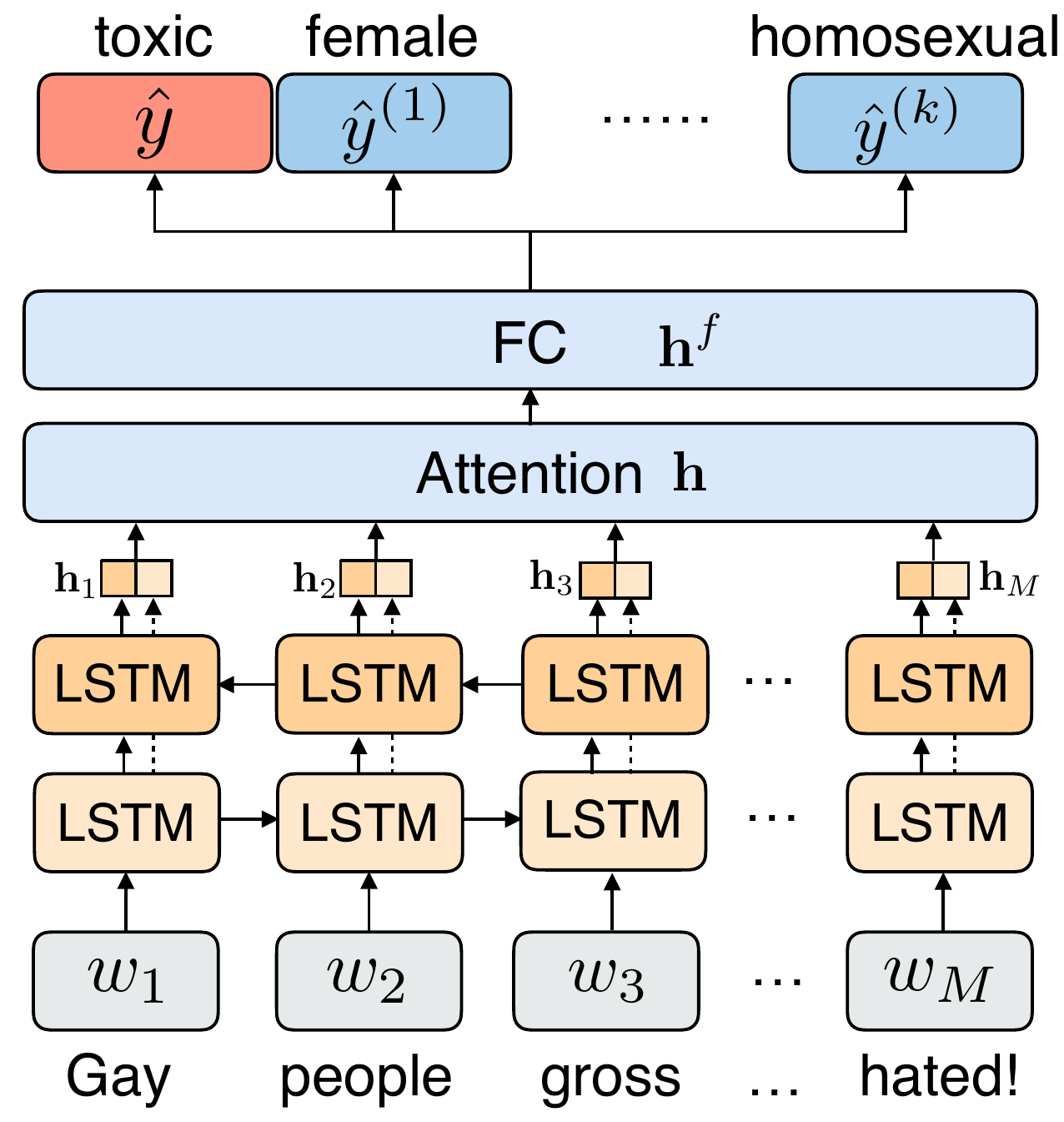}
     \caption{An overview of the proposed MTL frameworks. `FC' indicates a fully connected layer. Each label on the top represents a task.}\label{fig:overview}
  \end{figure}

\paragraph{Model}
The overview of the model is illustrated in Figure~\ref{fig:overview}. The highlights of our model include an embedding layer, two Long Short-Term Memory Network (LSTM) layers, an attention mechanism, and a custom loss function. Our multi-task learning model utilizes deep sharing to jointly learn the toxicity and identity tasks which allows us to exploit the commonalities and differences between tasks. The embedding layer allows our model to gain a better understanding of the semantics encoded within each word. The LSTM layers and attention mechanism work together to capture long-range and hidden-state dependencies, form a complete understanding of the entire document by parsing individual words, and pay specific attention to words that are relevant to each task. Finally, the custom-weighted loss function allows our model to place extra focus on learning to mitigate unintended bias, rather than simply increasing the ROC-AUC score for the entire test set. We are one of the first to implement a multi-task learning model that makes use of all of these components specifically to investigate the mitigation of unintended bias. Each of these elements is explored in more detail below.

\paragraph{Embedding \& LSTM layers }
Each word in a sentence is converted to a word embedding vector of dimension D which concatenates two parts: 1) pre-generated embeddings from the global vectors for word representation~\cite{pennington2014glove} and 2) pre-generated embeddings from the vectors provided by FastText~\cite{joulin2016bag}. Assuming there are $N$ total comments in the training dataset, each comment example has M words (M = max length) and is represented as $\mathbf{s}=[\mathbf{x}_1,...,\mathbf{x}_M]$. Each comment is associated with a toxicity label $y$ and a set of identity labels $y^1,...,y^K$ (K = number of identity labels). Each word $\mathbf{x}_m \in \mathbb{R}^D$ is represented by an embedding vector. Then we apply a bi-directional recurrent neural network (e.g. LSTM), a forward LSTM and a backward LSTM, to the sentence $\mathbf{s}$.  We obtain the hidden state $\mathbf{h}_m$ for each word $\mathbf{x}_m$ by concatenating the forward hidden state $\overrightarrow{\mathbf{h}}_m$ and the backward hidden state $\overleftarrow{\mathbf{h}}_m$. 

 \paragraph{Attention}
Attention mechanisms have shown to produce state-of-the-art results in many natural language processing tasks such as machine translation~\cite{attention:2015} when combined with neural word embeddings. In this paper, we explore a feed-forward attention mechanism~\cite{fnn-attention-16} on the bidirectional LSTM to ``memorize'' the influence of each hidden state:
% \begin{equation}
% \mathbf{u}_m= \tanh(\mathbf{W}_a \mathbf{h}_m) 
% \end{equation}
\begin{equation}
    a_m = \frac{\exp (\tanh(\mathbf{W}_a \mathbf{h}_m) )}{\sum_{j=1}^M \exp(\tanh(\mathbf{W}_a  \mathbf{h}_j) )} ,
\end{equation}
\begin{equation}
    \mathbf{h} =  \sum_{m} a_m \mathbf{h}_m ,
\end{equation}
where $a_m$ measures the importance of current word $m$ and $\mathbf{W}_a$ is the weight parameter to be learned.
Then two fully connected dense layers are applied on the hidden state of the sentence (comment) $\mathbf{h}$. This attention mechanism will allow our model to directly access the entire sequence. Our model will pay more ``attention'' to the words that correlate with toxicity and identity labels. Thus it helps to avoid classifying these comments without a deeper and more meaningful understanding of each document.

\paragraph{Multi-Task Learning}
Rather than learning each task individually, learning multiple tasks simultaneously has been theoretically and empirically proven to improve prediction performance~\cite{Caruana93multitasklearning}. Multi-task learning works the best when multiple tasks are related in some shape or form~\cite{mtl:2007}. In order to reduce unintended model bias, we take advantage of multi-task learning to model related tasks and capture their internal patterns. For instance, when predicting the toxic comment ``gay people are gross and universally hated'', the toxicity task will focus on the toxic elements ``gross and universally hated'' while the identity task will identify the trigger word ``gay'' in the comment. We expect involving identity tasks will reduce model bias by mitigating the confusion between identity and toxicity in predictions. Our model will utilize hard-parameter sharing as specified in~\cite{Ruder2017AnOO} which prevents the model from overfitting and allows for more opportunities to share information between the toxicity and identity classifications.

\paragraph{Prediction}
As shown in Figure~\ref{fig:overview}, for different tasks, we share the same structure of the network till the last output layer.
The predictions for toxic label $\hat{y}$ and identity labels $\hat{y}^k (k=1...K)$ are then modeled as below:

\begin{equation}\label{mtl:deep1}
  \hat{y} = \sigma (\mathbf{w}_t^\top\mathbf{h}^{f}+ b_t),
\end{equation}

\begin{equation}\label{mtl:deep2}
  \hat{y}^k = \sigma (\mathbf{w}_k^\top\mathbf{h}^{f}+ b_k), k = 1...K,
\end{equation}

where $\mathbf{h}^{f}$ is the output from the dense layers. We have also evaluated different levels of sharing. For instance, another design is sharing the bi-directional LSTM layers among tasks but each task has its own fully connected layers. We observe that the best performance comes from the deep sharing mechanism as demonstrated as Eq.~\ref{mtl:deep1} and~\ref{mtl:deep2}.

\paragraph{Model loss}
Finally, we design a weighted binary cross-entropy (CE) loss for all the label estimates of training examples. Given that a comment can have multiple identity labels, a general cross-entropy is not used in this case:
\begin{equation}
    L = \sum_{n=1}^N 
    \beta_n
    \big[
    \alpha 
    J_{\text{CE}} 
    (\hat{y}_n, y_n) 
    + (1-\alpha)
    \sum_{k=1}^K 
    J_{\text{CE}}  
    (\hat{y}^k_n, y^k_n)\big].
\end{equation}
We employ a weighted loss per example ($\beta_n$) and per task ($\alpha$). By default, $\beta_n=1$.  If an example is a non-toxic example with identity information, its weight $\beta_n = \beta_n \times c$ where c is a constant (e.g. $c=3$ in our experiments).  The task weight $\alpha \in [0,1]$ is selected by a grid search in validation set. 
All model parameters are trained via back-propagation and optimized by the Adam algorithm~\cite{kingma2014adam} given its efficiency and ability to avoid overfitting.

\section{Experimental Study}

The purpose of our experiment is to compare the performance of our multi-task learning model to other baseline models. The four types of baseline models used for comparison are: Logistic Regression, CNNs, LSTMs, and GRUs. Our hypothesis is that the multi-task learning model will be able to outperform the other baseline models in multiple categories, especially in those that measure unintended bias. In this experiment we focus on the toxicity task and the toxicity scores predicted by the model rather than the identity scores.

\subsection{Experiment Setup}
 \paragraph{Text preprocessing.} Before the model training, we perform some basic preprocessing on the data. To convert the raw text to a usable format, we first tokenize the comments. Because comments vary in length, the max-length is defined as 220 words. Sequences that had less than 220 words are padded with zeroes. During the process of tokenization, each comment is stripped of certain punctuation marks but was not converted to lowercase. 
 
\begin{figure}[h]
\centering
  \includegraphics[width=0.35\textwidth]{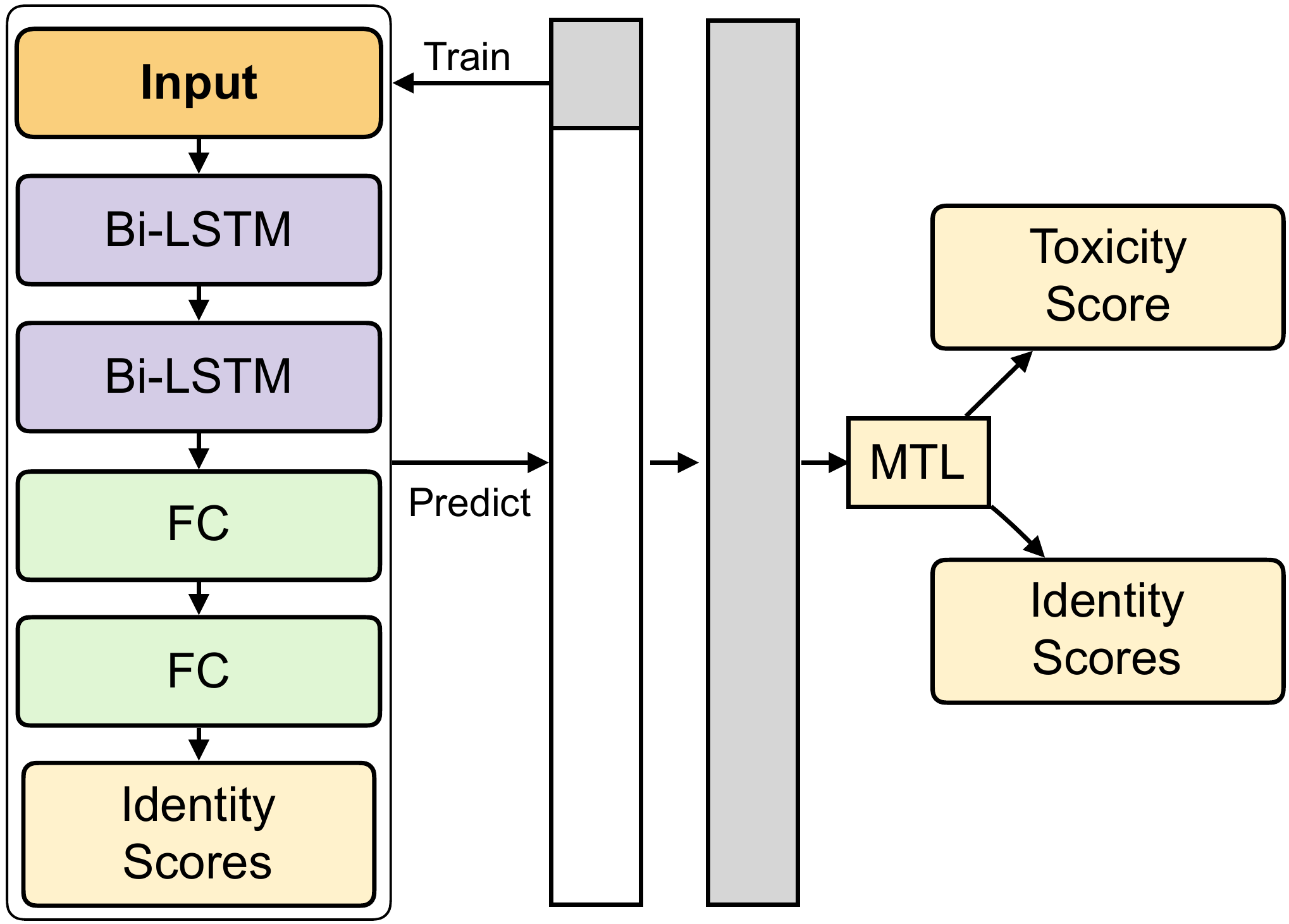}
     \caption{Multi-task learning model with model-specific preprocessing for propagating identity labels.}\label{fig:annotation}
  \end{figure}

  \paragraph{Model-specific preprocessing.} We also perform preprocessing specific to the multi-task learning model. Because only 405,130 out of 1,804,874 comments are annotated for each of the identities, we need to fill in the scores for the rest of the identities in order to employ an effective multi-task model. To fill in the rest of the identities, we train a multi-class classifier on the \char`\~400,000 training examples with the annotated identity scores to predict the identity scores for the remaining \char`\~1.4 million training examples. This multi-task model employs the same architecture as the MTL model that we discussed in the last section. However, we omit an attention layer from this model because we found that an attention layer does not significantly improve the accuracy of predicting the identities within a comment. The results from our MTL model is then fed into our multi-task learning model for prediction as shown in Figure~\ref{fig:annotation}.

  \noindent\paragraph{Cross-Validation} In our experiments, we perform K-fold (K=5) cross-validation on the dataset. In each fold, 80\% of the data is set aside for training and 20\% is used for validation, which translates into roughly 1.4 million and 400,000 comments respectively. 

 \noindent\paragraph{Model parameters and hyper-parameter settings} We select $\alpha$ (model loss eq.) by a grid search in our validation set and $\alpha$ is chosen to be $0.6$ with the best performance. We set the dimension of hidden states in the two bidirectional LSTM layers to be 256 and 512 for the two fully-connected layers. Rectified activation functions are applied to the fully-connected layers and a sigmoid activation function is applied to the output layers. We also introduce a spatial dropout of 20\% between the embedding and first bidirectional layer.
 
\subsection{Comparison Methods (Baseline Models)}

Each of the following baseline models was developed using the Keras framework. A significant portion of each of the following baseline models was adapted from existing works, albeit small changes in number and size of layers used.

\begin{itemize}
\item \textbf{Logistic Regression}
Logistic Regression~\cite{neter1996applied} has widely been used for binary classification tasks. For text classification tasks, documents are usually vectorized into bag-of-words (BoW) features (e.g. TF-IDF). 
As a comparison to dense vectors in deep learning models, our model applies a TF-IDF vectorizer to the raw comments and then passes it through a standard logistic regression model to obtain the final predictions. 

\item \textbf{Convolutional Neural Networks}
Convolutional Neural Networks~\cite{lecun:1998} have proved to be very successful when it comes to sentence or character-level sentence classification~\cite{kim-2014-convolutional}. CNNs have been known to work better for datasets with a large amount of training examples and can work well for user-generated data, given its ability to deal with the ``obfuscation of words'' in comments and ``detect specific combinations of features'' in text classification. Our CNN model is adapted from the following paper~\cite{Georgakopoulos:2018:CNN}.

\item \textbf{Long Short-Term Memory Network}
LSTMs~\cite{Hochreiter:1997:LSM} were introduced primarily to overcome the problem of the vanishing gradient. As a variant of Recurrent Neural Networks (RNN), 
it has proven to have a better ability to learn long-range dependencies. In the Simple LSTM baseline model, we introduced a 20\% spatial dropout. The input is passed through two LSTM layers of 256 units each. Afterwards, the input passes through two dense layers of 512 units each with a rectified linear activation function. Finally, we obtain a single output (toxicity score) by applying a sigmoid activation function to the final dense layer. The architecture for LSTMs and GRUs were adapted from the following paper~\cite{van-aken-etal-2018-challenges}.

\item \textbf{Gated Recurrent Unit}
GRU~\cite{chung:2014} operates similarly to an LSTM but instead uses a reset and update gate, where the reset gate acts to forget the previous state and the update gate decides how much of the candidate activation to use in updating the cell state. Our GRU model is similar to the structure of our LSTM model, with the exception that only 128 units were used per GRU layer.

\item \textbf{Bidirectionality}
Introducing bidirectionality into a RNN can help a network learn from both past and future context~\cite{Schuster:1997:BRN}. 
In this architecture, two layers of hidden nodes are introduced. In the second layer, the input is reversed and the sequence is passed backwards into the network. Within the scope of this task, understanding and learning from a sequence in both directions can lead to a more complex and more accurate understanding of the document. In this paper, we implement a Bidirectional LSTM and a Bidirectional GRU. They follow the same structures as the Simple LSTM and GRU specified in the previous paragraphs. 

\item \textbf{MTL-Aux}
In addition to the baseline models, we developed another multi-task learning model for comparison. Instead of using the nine identity labels, MTL-Aux  focuses on five alternate subtype toxicity attributes: \textit{severe toxicity}, \textit{obscene}, \textit{threat}, \textit{insult}, and \textit{identity attack}.
The model follows the exact same structure as MTL-attn but predicts the aforementioned five subtype outputs rather than the nine identity outputs. 
\end{itemize}

\subsection{Evaluation Metrics}
For basic toxicity detection evaluation, we calculate the ROC-AUC, precision, recall, and F1-scores for the full test set. Predictions with toxicity scores that are greater than or equal to 0.5 are considered to be part of the positive (toxic) class and vice-versa.

Unintended bias evaluation metrics are introduced and specified by the Google Conversation AI Team in their paper~\cite{Borkan:2019:NMM}.  The Generalized Mean of Bias AUCs metric was introduced by their Kaggle competition\fnref{jigsaw}. The following evaluation metrics are specifically crafted to accurately measure the reduction of unintended bias by a model. By restricting the test set, we get a better understanding of how each of the models perform within the scope of toxic comment detection and bias reduction.
\begin{itemize}

\item \textbf{Subgroup AUC} We restrict the test set to comments for which each identity label is positive. An ROC-AUC score is then calculated for each of the identity groups which is hereby called the Subgroup AUC. A low value indicates that the model does a bad job of distinguishing between toxic and non-toxic comments in the context of that specific identity (e.g. gay, muslim, black). 

\item \textbf{BPSN (Background Positive, Subgroup Negative) AUC} To calculate this metric, we restrict the test set to non-toxic comments that mention the identity and toxic comments that don't mention the identity. We obtain the BPSN AUC by getting the ROC-AUC score from this restricted test set. The main purpose of this metric is to measure the false positive rate of each model in the context of each specific identity. A higher BPSN AUC score means that a model is less likely to confuse non-toxic examples that mention the identity with toxic examples that do not, meaning that the model is able to mitigate biases towards a specific identity. BPSN AUC can be considered a stronger evaluation metric than Subgroup AUC because it tailors towards the specific focus of this paper: reduce the false positive rate towards certain identities.

 \item \textbf{Generalized Mean of Bias AUCs}
 One overall measure is calculated from the Subgroup AUCs using the following formula: 
$M_p(m_s) = (\frac{1}{N}\sum_{s=1}^{N}m^p_s)$
\noindent
where $M_p$ is the $p$-th power-mean function, $m_s$ is the bias metric calculated for subgroup $s$, and $N=9$ which is the number of identity subgroups. We set $p=-5$ as suggested in the competition. A low value indicates model bias toward one or more of the  identities. This metric is essentially a average of all nine subgroup AUCs. This metric is hereby referred to as Generalized Mean AUC throughout the rest of the paper.

\end{itemize}

\section{Results \& Empirical Analysis}

\begin{table*}[t]
\def\arraystretch{1.1}
\centering
\caption{Binary classification performance of different methods on toxic comments. Bold face indicates the best result of each column and underlined the second-best. Codes are used in subsequent tables to refer to each of the models.  }\label{tab:overall-result}
\begin{threeparttable}
\begin{tabularx}{0.78\pdfpagewidth}{l c c c c c c}
\toprule
\textbf{Model}                 & \textbf{Generalized Mean AUC}  & \textbf{AUC}   & \textbf{Precision} &\textbf{Recall} &\textbf{F1-Score} & \textbf{Code}\\
\hline
Logistic Regression         &0.8999   &0.9488	&0.79	&0.50	&0.61	&LR \\
CNN                         &0.9212   &0.9635	&\underline{0.86}	&0.47	&0.60  &CNN  \\
Simple LSTM                 &0.9267   &0.9662	&0.85 &0.50	&0.63	 &LSTM\\
Bidirectional LSTM          &0.9316   &0.9694	&0.83	&0.55	&\underline{0.66} &Bi-LSTM	 \\             
Bidirectional LSTM \& Attn  &0.9322   &0.9696	&0.84	&\underline{0.55}	&\underline{0.66} &Bi-LSTM-A	 \\  
Simple GRU                  &0.9284   &0.9676	&0.83	&0.54	&0.65 &GRU	 \\               
Bidirectional GRU           &0.9319   &0.9637	&0.84	&0.52	&0.64 &Bi-GRU	 \\              
Bidirectional GRU \& Attn   &\underline{0.9325}   &\underline{0.9697}	&0.84	&0.53	&\underline{0.66} &Bi-GRU-A	 \\  
MTL-Aux                     &0.9317   &0.9693	&\underline{0.86}	&0.53	&0.65 &MTL-Aux	 \\  
\hline
MTL-attention                 &\textbf{0.9407}$^\star$  &\textbf{0.9709}	&\textbf{0.88}$^\star$ 	&\textbf{0.59}$^\star$	&\textbf{0.71}$^\star$ & MTL-attn	 \\
\bottomrule
\end{tabularx}
\begin{tablenotes}
      \item $^\star$ Identifies statistical significance ($p<0.05$) compared to best baseline model in the category. 
    \end{tablenotes}
    \end{threeparttable}
\end{table*}

\begin{figure*}
\centering
\centering
  \includegraphics[width=0.7\textwidth]{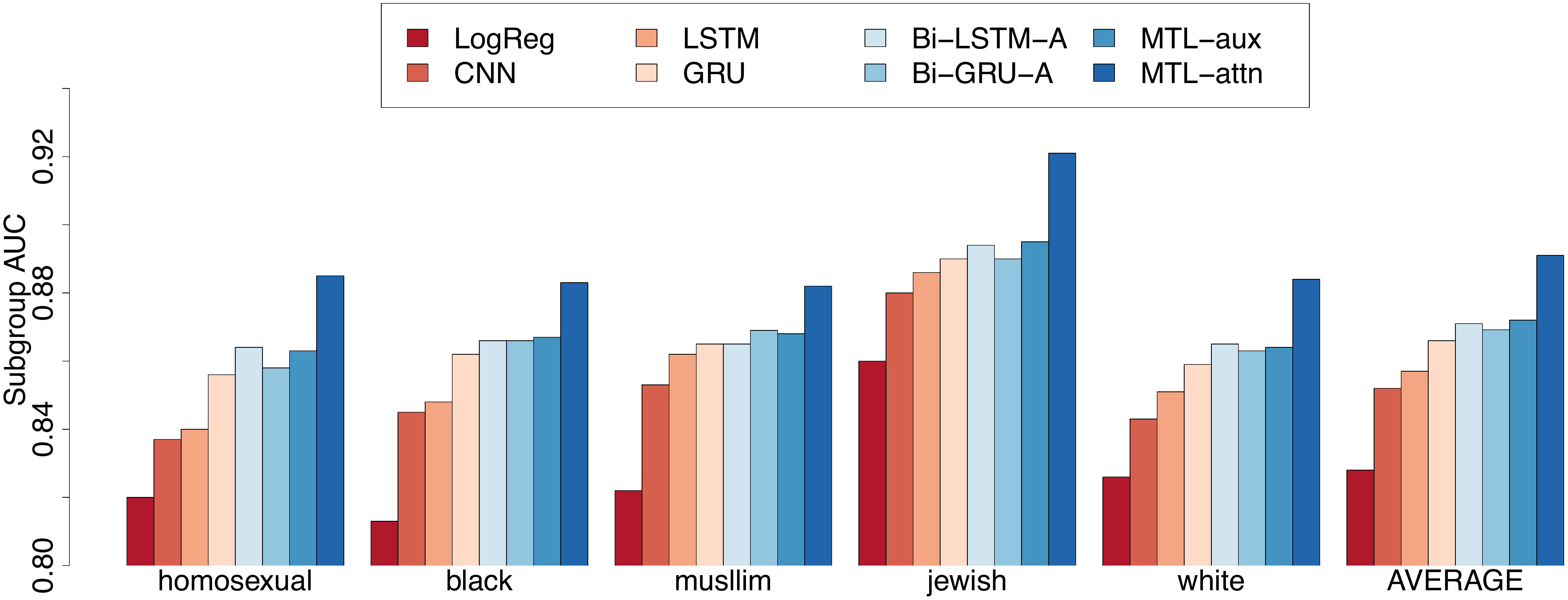}
  \caption{Subgroup AUC}\label{fig:subauc}
\centering
  \includegraphics[width=0.7\textwidth]{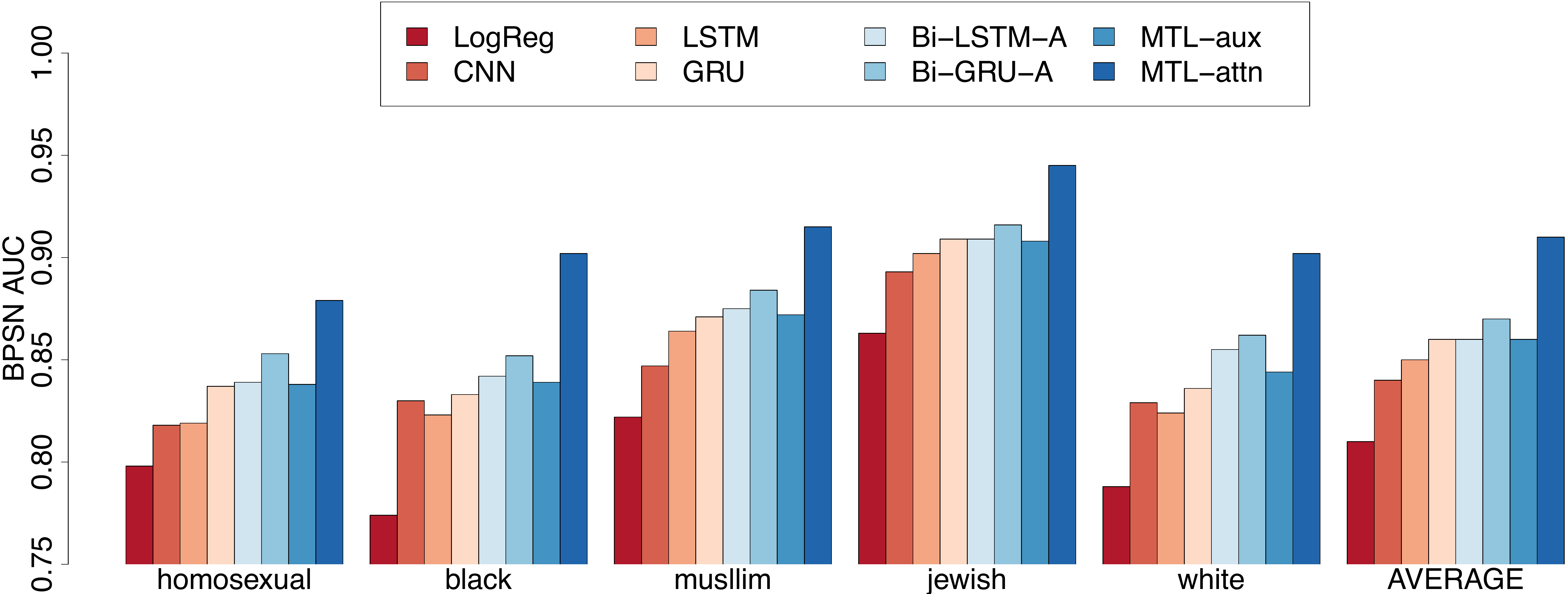}
  \caption{BPSN AUC}
 \label{fig:bpsnauc}
  \end{figure*}

\subsection{Proposed Model: MTL-attention}
The overall binary classification performance is summarized in Table \ref{tab:overall-result}. The MTL-attention model outperforms all other baseline models in Generalized Mean Bias AUC, suggesting that it was the most successful at accurately classifying comments with any of the aforementioned identities present. MTL-attention also outperformed all other models in recall, precision, and F1-score. This strongly suggests that learning tasks in parallel could be useful in forming a shared representation of the dataset, where what is learned for one task helps other tasks to be learned better. In addition, the inclusion of the attention layer and custom loss function enabled the model to pay closer attention to certain words and phrases that signaled the possibility of a false positive classification. 

Our results are statistically significant when compared to the best baseline model. We conduct a Kolmorogov-Smirnov test to evaluate the difference in means for non-Gaussian results. 
% Because K=5 in our K-Fold cross validation of the dataset, our sample size was 5. 
For Generalized Mean AUC, precision, recall, and F1-score, we found the p-value to be below 0.05 which implies that the improvement in performance between MTL-attention and the best performing baseline model was statistically significant for these evaluation metrics. From these results, we conclude that our multi-task learning model introduces a low level of variance and is effective at reducing unintended model bias when compared to current state-of-the-art models.

While an improvement of 0.8\% in the Generalized Mean AUC may seem incremental, it is in fact significant when observing its effect on the rate of false positives. With approximately 191,671 comments being labeled as being associated with an identity and about 14.44\% being non-toxic, we have 27,677 non-toxic comments that are associated with an identity. If the main improvement achieved by the multi-task learning model was in reducing bias and the false positive rate (evidenced by an increase in precision), then a 0.8\% improvement in Generalized Mean AUC can imply a significant improvement for non-toxic comments. This idea is further explored within the following Subgroup \& BPSN AUC subsection. 

\subsection{Baseline Models}
The two best baseline models are Bi-LSTM-A and Bi-GRU-A with the highest Generalized Mean AUCs and the highest F1-scores. Bi-LSTM and Bi-GRU follow close behind, suggesting that an attention layer by itself does not contribute significantly to an improvement in model performance. MTL-Aux had comparable performance to Bi-LSTM, suggesting that multi-task learning with the five subtype attributes (\textit{severe toxicity}, \textit{obscene}, \textit{threat}, \textit{insult}, \textit{identity attack}) does not serve to improve performance in the test set. This is understandable given that four subtypes are not significantly related to identity information and that \textit{identity attack} does not recognize different identity groups.

CNN outperforms all other models in precision despite having the lowest overall F1-score. We believe the convolutional layer helps in capturing key local patterns with respect to the toxicity score. We also observe an extremely strong showing in performance from LSTMs and GRUs, which is primarily due to their ability to retain memory, helping solve problems related to long-range dependencies. Introducing bidirectionality to LSTMs and GRUs also offers a significant advantage in performance because the model has the access to the entire context of a comment by having it passed in forwards and backwards. It is, therefore, able to gain a better understanding of the entire document and able to parse individual words into coherent and understandable utterances. Logistic Regression has the lowest performance thus far, which can be explained by the sparse nature of a Bag-of-Words model and the disregard of the order of words in a sentence.
\begin{table*}[t]
    \centering
    \small
      \caption{Templates of non-toxic and toxic comments with identity labels. (Comment Comparison Case Study)}
    \label{tab:templates}
    \begin{tabular}{p{8cm}p{8cm}}
    \toprule
    \textbf{Non-Toxic templates(\textcolor{green}{\textbf{-}})} & \textbf{Toxic templates(\textcolor{red}{\textbf{+}})}\\
    \midrule
     (1). \textit{I am $\langle$Identity$\rangle$} &  (1).  \textit{I hate all $\langle$Identity$\rangle$}\\
     (2). \textit{I am a $\langle$Identity$\rangle$ person, ask me anything} &(2). \textit{I am a $\langle$Identity$\rangle$ person and I hate your guts and you suck}\\
      (3). \textit{$\langle$Identity$\rangle$ people are just like everyone else} &  (3).   \textit{$\langle$Identity$\rangle$ people are gross and universally hated}\\
    (4). \textit{I am friends with a $\langle$Identity$\rangle$ person} & (4).   \textit{I disagree with you because you are $\langle$Identity$\rangle$}\\
          (5).    \textit{Why should $\langle$Identity$\rangle$ people be treated any differently?} &(5).     \textit{F**k all $\langle$Identity$\rangle$ people!}\\
       (6).    \textit{I hate when $\langle$Identity$\rangle$ people are stereotyped} & (6).  \textit{$\langle$Identity$\rangle$ people are anti-god}\\
         (7).    \textit{I have no opinion on $\langle$Identity$\rangle$ people} &(7).      \textit{I'm going to kill all $\langle$Identity$\rangle$ people one day}\\
        \bottomrule
    \end{tabular}
\end{table*}

\subsection{Subgroup \& BPSN AUCs}

As show in Figures~\ref{fig:subauc} and~\ref{fig:bpsnauc}, MTL-attention significantly outperforms other models in both Subgroup AUC and BPSN AUC. The focus of this paper from the beginning has been to observe the ability of multi-task learning to mitigate bias for the following identities: \textit{homosexual}, \textit{black}, \textit{muslim}, and \textit{jewish}. In Figure~\ref{fig:subauc}, we observe a 3-5\% increase in Subgroup AUC for each of the aforementioned categories when compared to the best baseline model and an average improvement of around 3\%. The next best baseline models are generally equivalent in performance to each other, being the LSTM, GRU, and their bidirectional counterparts.

In the context of the BPSN AUC metric (Figure~\ref{fig:bpsnauc}), significant improvements were also noted for the aforementioned identities. On average, the MTL-Attn model achieves an increased BPSN AUC performance of approximately 5-7\%, suggesting it is able to considerably reduce the false positive rate by jointly learning the identity and toxicity tasks. Overall, the results show that our multi-task learning model was able to achieve its goal of improving performance for identities that are historically and empirically found to introduce bias into a model. This improvement in Subgroup AUC and BPSN AUC further solidifies that the improvement of 0.8\% in the Generalized Mean AUC correlated primarily with a mitigation of unintended bias and the false positive rate for the aformentioned identities.

\section{Case Study: Comment Comparison}

The main goal of this case study is to test our multi-task learning model on a variety of real-life comments that our model could encounter within online conversations on social media platforms. This experiment takes a look at the toxicity scores given individual toxic and non-toxic comments that make reference to certain identities. Two sets of templates with seven comments each are selected to simulate potential comments on social media platforms as shown in Table~\ref{tab:templates}.  We follow Dixon et al.~(\citeyear{Dixon:2018:MMU}) to sample the first three comments in each template. The rest of the non-toxic comments were randomly sampled within the dataset for comments under 50 characters. The rest of the toxic comments were created to represent one of each of the five subtype attributes previously mentioned: \textit{obscene}, \textit{threat}, \textit{sexually\_explicit}, and \textit{insult}. For each set of templates (non-toxic and toxic), $\langle$Identity$\rangle$ is replaced by one of the following six keywords, meant to represent the identities defined in Table~\ref{tab:data}: \textit{gay}, \textit{lesbian}, \textit{bisexual}, \textit{muslim}, \textit{jew}, \textit{black}. Each comment is then tokenized and passed through each of the eight models specified below to obtain a toxicity score (seven toxicity scores for non-toxic templates and seven toxicity scores for toxic templates per model). 
We obtain toxicity scores from Google's Perspective API framework with the intention of comparing it with our multi-task learning and baseline models. 
An example of a comment that we use to test model bias is ``I am a gay person, ask me anything.'' 
If a model gives a high toxicity score for this comment, the model is potentially biased towards the gay identity. Models with lower average non-toxic scores are less likely to be biased towards a specific identity. However, we still want to observe high scores for toxic comments because low scores for toxic comments may indicate that models are  unable to correctly distinguish between non-toxic and toxic comments and are simply ``shifting the decision boundary.''

\begin{figure}
    \centering
    \includegraphics[width=0.45\textwidth]{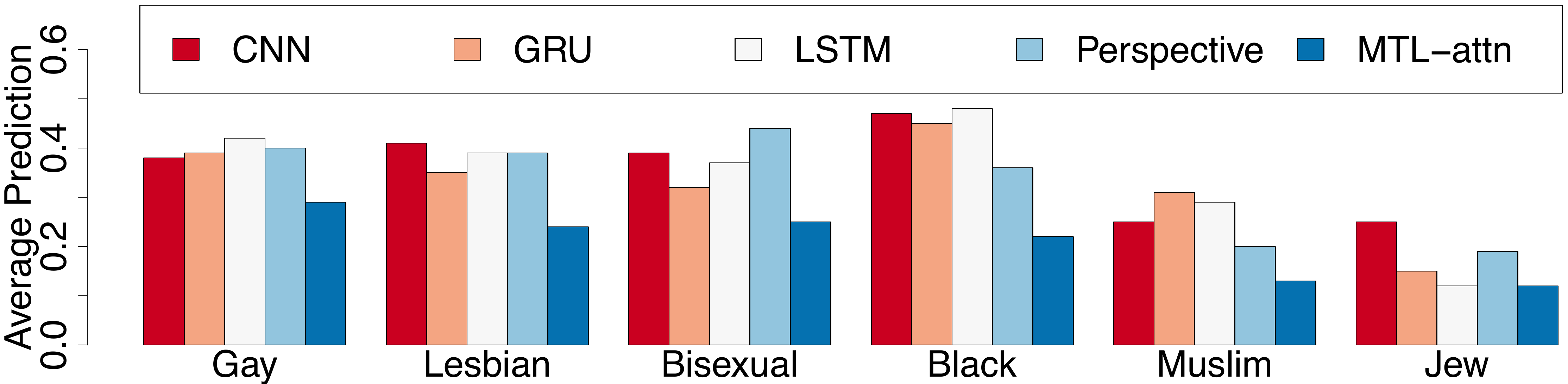}
    \caption{Average prediction scores on non-toxic templates for each identity group.}
    \label{fig:nontoxic}
     \includegraphics[width=0.45\textwidth]{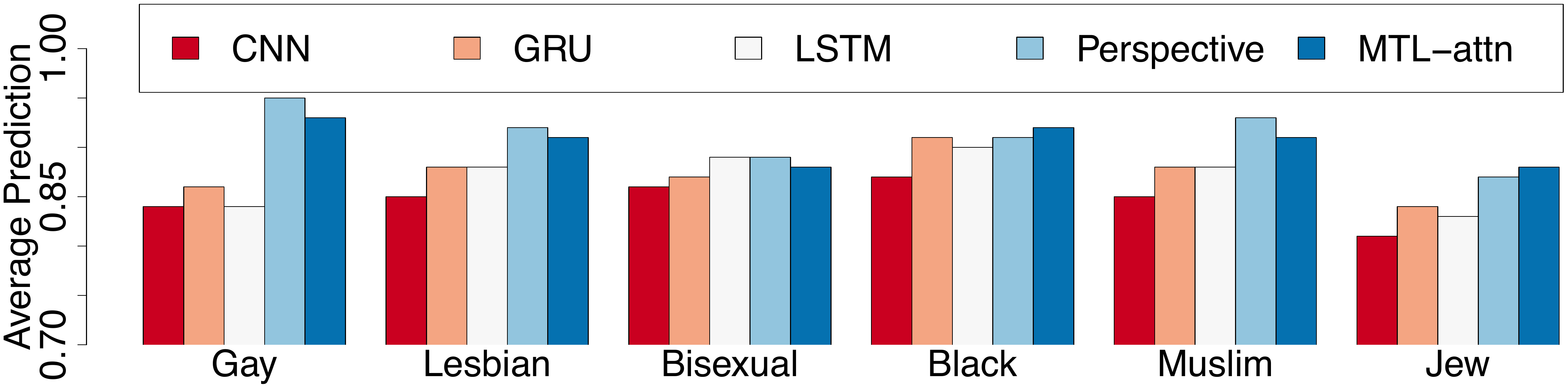}
    \caption{Average prediction scores on toxic templates for each identity group.}
    \label{fig:toxic}
\end{figure}
As shown in Figures~\ref{fig:nontoxic} and~\ref{fig:toxic}, we observe that multi-task learning models are able to distinguish very well between toxic and non-toxic comments for each of the identities. We see that our proposed model is consistently able to predict low scores for non-toxic comments and high scores for toxic comments. Within the non-toxic comments, we see significant improvements between the baseline models (including Perspective API) and our multi-task learning model. On average, our model predicts 10\% less toxicity for non-toxic comments that mention one of the aforementioned identities. For non-toxic comments, our model never predicts a toxicity score above 0.40 and, for toxic comments, it never predicts a score below 0.80. While Google's Perspective API does not misclassify any comments, it is important to note that their toxicity scores for gay, lesbian, and bisexual keywords are higher than expected. This suggests that there may be some bias towards these three identities which could potentially lead to problems in the future. The baseline models (CNN and LSTM) do not perform well, as evidenced by numerous misclassifications especially for the gay and black identities. 

\section{Conclusion \& Future Work}
In this work, we present an attention-based multi-task learning approach to reduce unintended model biases towards commonly-attacked identities in the scope of toxic comment detection. 
The proposed model outperformed other models in terms of metrics that specifically measure unintended bias, while still being able to correctly identify and classify toxic comments. 
Through our research, we noted that our multi-task learning models significantly improved classification performance when the comments are related to the following identities: \textit{homosexual}, \textit{muslim}, \textit{black}, and \textit{jewish}. We also conducted a case study which demonstrated the robustness of our model and its ability to perform well within a variety of situations. Overall, the empirical results confirm our initial hypothesis that learning multiple related tasks simultaneously can bring advantages to reducing biases towards certain identities while improving the health of online conversations. 
However, the proposed method is limited in terms of semantic encoding abilities and the model is not flexible when new or hidden identities appear.

In the future, we plan to focus on a few directions: 1) Given the limited identity labels for comments, we will explore pre-trained models such as semi-supervised knowledge transfer models or BERT. 2) We will also investigate other hidden cultural bias in online toxic comments other than identities. 3) Most existing models focus on the prediction of toxic comments and identity group recognition. We will study interpretable machine learning methods to identity the trigger words or phrases for determining toxicity in a comment. 

\begin{footnotesize}
\bibliographystyle{aaai}
\bibliography{ref}
\end{footnotesize}%

\end{document}